\documentclass[10pt, a4paper, conference, compsocconf]{IEEEtran}

\usepackage{cite}
\usepackage{amsmath,amssymb,amsfonts}
\usepackage{graphicx}
\usepackage{textcomp}
\usepackage{color}
\usepackage{url}
\usepackage{booktabs}
\usepackage{makecell}
\usepackage[flushleft]{threeparttable}
\usepackage{multirow}

\begin{document}
%
\title{The Robust Reading Competition Annotation and Evaluation Platform}


\author{
	\IEEEauthorblockN{
		Dimosthenis Karatzas, 
		Lluis G\'omez,
		Anguelos Nicolaou,
		Mar\c{c}al Rusi{\~n}ol
	} \\
	
	\IEEEauthorblockA{\textit{Computer Vision Centre, Universitat Autonoma de Barcelona, Barcelona, Spain;}}
	\IEEEauthorblockA{\textit{\{dimos, lgomez, anguelos, marcal\}@cvc.uab.es}}
}


\maketitle

\begin{abstract}
The ICDAR Robust Reading Competition (RRC), initiated in 2003 and re-established in 2011, has become a de-facto evaluation standard for robust reading systems and algorithms. Concurrent with its second incarnation in 2011, a continuous effort started to develop an on-line framework to facilitate the hosting and management of competitions.

This paper outlines the Robust Reading Competition Annotation and Evaluation Platform, the backbone of the competitions. The RRC Annotation and Evaluation Platform is a modular framework, fully accessible through on-line interfaces.
It comprises a collection of tools and services for managing all processes involved with defining and evaluating a research task, from dataset definition to annotation management, evaluation specification and results analysis.

Although the framework has been designed with robust reading research in mind, many of the provided tools are generic by design.
All aspects of the RRC Annotation and Evaluation Framework are available for research use.

\end{abstract}


%
\IEEEpeerreviewmaketitle

\section{Introduction}
The Robust Reading Competition (RRC) series\footnote{\url{http://rrc.cvc.uab.es/}} addresses the need to quantify and track progress in the domain of text extraction from a variety of text containers like born-digital images, real scenes, and videos. The competition was initiated in 2003 by S.Lucas et al.~\cite{lucas2003icdar} initially focusing only on scene text detection and recognition, and was later extended to include challenges on born-digital images~\cite{karatzas2011icdar}, video sequences~\cite{karatzas2013icdar}, and incidental scene text~\cite{karatzas2015icdar}. The $2017$ edition of the Competition introduced five new challenges on: scene text detection and recognition based on the COCO-Text dataset~\cite{gomez2017DOST}; text extraction from biomedical literature figures based on the DeText dataset~\cite{yang2017DeText}; video scene text localization and recognition on the Downtown Osaka Scene Text (DOST) dataset~\cite{iwamura2017DOST}; constrained real world end-to-end scene-text understanding based on the $>1M$ images French Street Name Signs (FSNS) dataset~\cite{smith2016end};  Multi-lingual scene text detection and script identification~\cite{nibal2017MLT}; and information extraction in historical handwritten records~\cite{fornes2017IEHHR}.

To manage all the above Challenges and respond to the increasing demand, we have invested significant resources to the development of the RRC Annotation and Evaluation Platform, which is the backbone of the competition.

Our goals, while working on the RRC platform were (1) to define widely accepted, stable, public, evaluation standards for the international community, (2) to offer open, qualitative and quantitative evaluation and analysis tools, and (3) to register the evolution of robust reading research, acting as a virtual archive for submitted results.

\begin{figure}[!t]
	\centering
	\includegraphics[width=\linewidth]{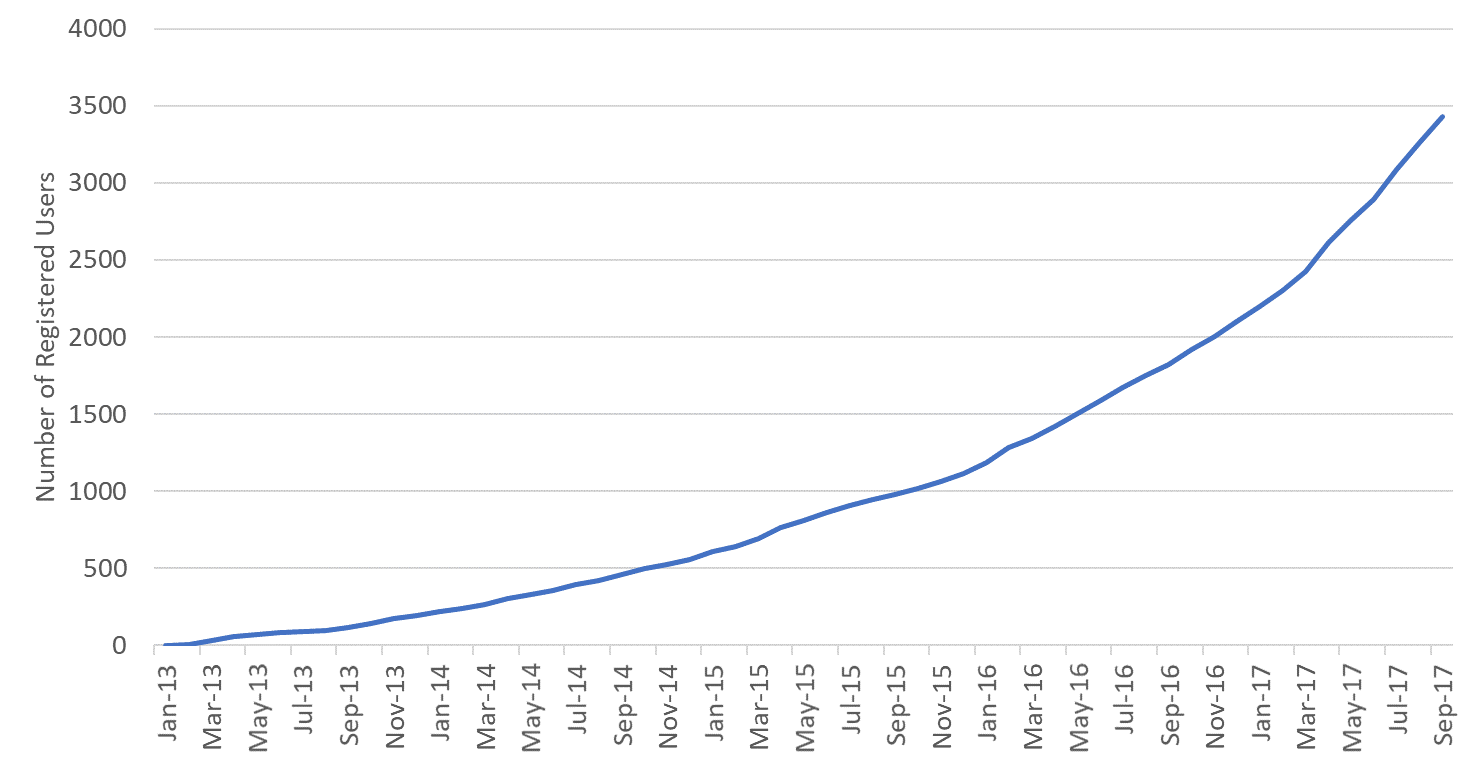}
	\caption{The evolution of registered users since.}
	\label{fig:UserEvolution}
\end{figure}

Supported by the evolving RRC platform, the competition has steadily grown, and the platform itself has been exposed to real-life stress. Over the past four years the RRC Web portal has received $>500,000$ page views\footnote{Measured with Google Analytics}. At the time of writing, the competition portal has more than $4,100$ registered users from more than $90$ countries. The evolution of registered users has been exponential, currently receiving $~10$ new registration requests per day (see Figure \ref{fig:UserEvolution}).

Registered researchers have submitted to date more than $15,000$ results that have been automatically evaluated on-line using the platform's afforded functionality. In many cases, the Web portal is used as a research tool by researchers who log their progress by consistently evaluating and comparing their results to the state of the art. Consequently, the portal receives and evaluates on average $20-30$ new submissions per day, while most of the evaluations are kept private by their authors as a private log. Out of the submitted methods, $553$ have been made public. A summary of the submissions received by the time of writing is given in Table~\ref{tab:RRC_submissions}.

\begin{table}
	\caption{Number of submissions to the different RRC challenges.}
	\label{tab:RRC_submissions}
	\begin{threeparttable}
		\centering
		\begin{tabular*}{\linewidth}{l @{\extracolsep{\fill}} ccl}
			\toprule
			& \makecell{Public\\Submissions} & \makecell{Private\\Submissions} & \makecell{Years\\Active} \\
			\midrule
			Born Digital			& 66	& 1,443	& 2011 - 2017\\
			Focused Scene Text		& 155	& 7,228	& 2003 - 2017\\
			Text in Video			& 18	& 436	& 2013 - 2017\\
			Incidental Scene Text	& 122	& 2,571	& 2015 - 2017\\
			COCO-Text				& 35	& 241	& 2017\\
			FSNS					& 1		& 0		& 2017\\
			DOST					& 14	& 0		& 2017\\
			MLT						& 107	& 145	& 2017\\
			DeText					& 20	& 76	& 2017\\
			IEHHR					& 15	& 2		& 2017\\
			\bottomrule
		\end{tabular*}
		\begin{tablenotes}
			\item[] \textit{Data valid on December 2017.}
		\end{tablenotes}
	\end{threeparttable}
\end{table}

Behind the scenes, the set of tools and services that constitute the RRC Annotation and Evaluation Platform is what has made it possible to scale in such a significant rate, and keep pace with new demands. This paper describes certain aspects of the platform, and indicates how different functionality is made available to researchers through a variety of software tools and interfaces.

\section{Background}
The need for reproducible research is a long standing challenge. Recently, workshops like RRPR 2016\footnote{1st Int. W. on Reproducible Research in Pattern Recognition} and OST 2017\footnote{1st Int. W. on Open Services and Tools for Document Analysis} have highlighted the challenge, while European research policies on Open Science and Responsible Research and Innovation promote related actions. Achieving truly reproducible research is a multifaceted objective to which research communities as much as individuals have to commit to, and involves among others curating and publishing data, standardising evaluation schemes, sharing code, etc.

Open platforms that aim to facilitate one or more of these aspects are currently available, ranging from the EU's catch-all repository Zenodo\footnote{\url{http://zenodo.org}} to GitHub\footnote{\url{https://github.com/}} for code sharing and Kaggle\footnote{\url{https://www.kaggle.com/}} for hosting research contests. Such platforms have witnessed a rapid growth and increased adoption by the research community over the past decade.

In the particular domain of document image analysis, open tools and platforms for research are a recurrent theme, including over the years attempts like the Pink Panther~\cite{yanikoglu1998pink}, TrueViz~\cite{lee2003architecture}, PerfectDoc~\cite{yacoub2005perfectdoc}, PixLabeler~\cite{saund2009pixlabeler}, PETS~\cite{seo2010performance} and  Aletheia~\cite{clausner2011aletheia},\cite{antonacopoulos2006ground}, to mention just a few.


In terms of more generic frameworks, the Document Annotation and Exploitation (DAE)\footnote{\url{http://dae.cse.lehigh.edu/DAE/}} platform~\cite{lamiroy2012non},\cite{lamiroy2016dae} consists of a repository for document images, implementations of algorithms and their results when applied to data in the repository. DAE promotes the idea of algorithms as Web services. Notably, it has been running since 2010 and is the preferable archiving system for datasets of IAPR-TC10. 

The more recent DIVAServices framework~
\cite{wursch2016divaservices}, is retake on the DAE idea, using a RESTful Web service architecture.

The ScriptNet platform\footnote{\url{https://scriptnet.iit.demokritos.gr/competitions/}}, developed through the READ project, offers another framework for hosting competitions related to Handwritten Text Recognition and other Document Image Analysis areas, and has been used to date for organising six ICDAR and ICFHR competitions.

What makes the RRC platform stand out is probably its large-scale adoption by the international research community. It should also be noted, that contrary to other initiatives, it was not originally conceived as a fully-fledged platform, but it has evolved responding to the needs of a particular research community as we perceived them over the years~\cite{karatzas2014online}. As such, it started off as a very specific set of tools aimed to help competition organisers in the particular field of robust reading, and has evolved into a much more generic platform, with an open set of tools and interfaces, covering all necessities related to defining, evaluating and tracking performance on a given research task.

Still, we do not perceive the RRC platform as a contestant to other initiatives, but rather as a useful contribution to a growing ecosystem of solutions. The present paper aims to shed some light behind the scenes of the Robust Reading Competition, by offering details about the platform that supports it and certain insights gained through using it over the years.


\section{The RRC Annotation and Evaluation Platform}
The RRC Annotation and Evaluation Platform, is a collection of tools and services, that aim to facilitate the generation and management of data, the annotation process, the definition of performance evaluation metrics for different research tasks and the visualisation and analysis of results. All on-line software tools are implemented as HTML5 interfaces, while specialised processing (e.g. the calculation of performance evaluation metrics) is based on Python and takes place on the server side. A summary of key functionalities of the platform is given in Table~\ref{tab:RRC_overview}.

\begin{table}
	\caption{Overview of functionality blocks of the RRC platform.}
	\label{tab:RRC_overview}
	\begin{threeparttable}
		\centering
		\begin{tabular*}{\linewidth}{l @{\extracolsep{\fill}} l}
			\toprule
			\multirowcell{2}{Dataset\\Management} & Data import \\
			\cmidrule{2-2}
			& \textbf{Specialised crawlers} \\
			\midrule
			\multirowcell{3}{Image\\Annotation} & \textbf{Annotation dashboard} \\
			\cmidrule{2-2}
			& \textbf{Annotation tools} \\
			\cmidrule{2-2}
			& \textbf{Quality control tools} \\
			\midrule
			\multirowcell{3}{Definition of\\Research Tasks} & Definition of subsets \\
			\cmidrule{2-2}
			& \textbf{Evaluation scripts} \\
			\cmidrule{2-2}
			& \textbf{Packaging and deployment} \\
			\midrule
			\multirowcell{4}{Evaluation and\\Visualisation\\of Results} & User and submissions management \\
			\cmidrule{2-2}
			& Uploading of results and automatic evaluation \\
			\cmidrule{2-2}
			& \textbf{Visualisation – getting insight} \\
			\cmidrule{2-2}
			& \textbf{Downloadable, standalone evaluation interfaces} \\
			\bottomrule
		\end{tabular*}
		\begin{tablenotes}
			\item[] \textit{In bold the functionalities discussed in more detail in this paper.}
		\end{tablenotes}
	\end{threeparttable}
\end{table}


\subsection{Dataset Management}
\begin{figure}[!t]
	\centering
	\includegraphics[width=\linewidth]{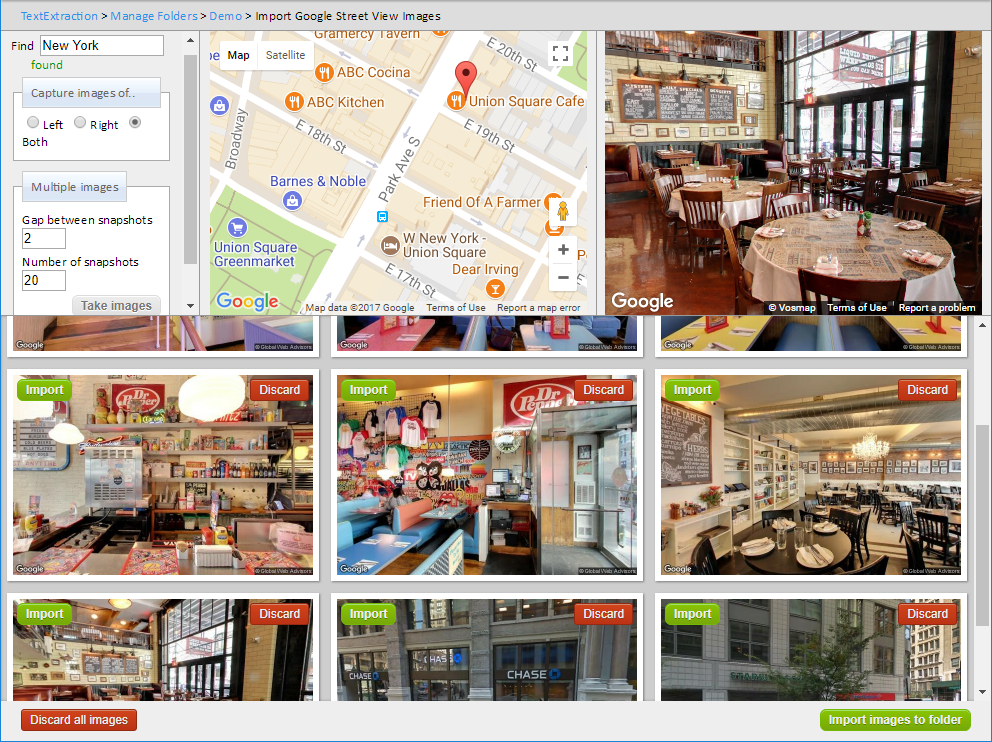}
	\caption{The integrated Street View crawler.}
	\label{fig:StreetViewCrawler}
\end{figure}

Datasets of images can be managed through Web interfaces supporting the direct uploading of images to the RRC server, but also offering tools to harvest images on-line. As an example, a Street View crawler is integrated in the RRC platform and can be used to automatically harvest images from Street View as seen in Figure~\ref{fig:StreetViewCrawler}.

The datasets are treated as separate collections, for which different levels of access can be defined for administrators, data owners and other contributors (e.g. annotators) to the collection.

\subsection{Image Annotation}
\begin{figure}[!t]
	\centering
	\includegraphics[width=\linewidth]{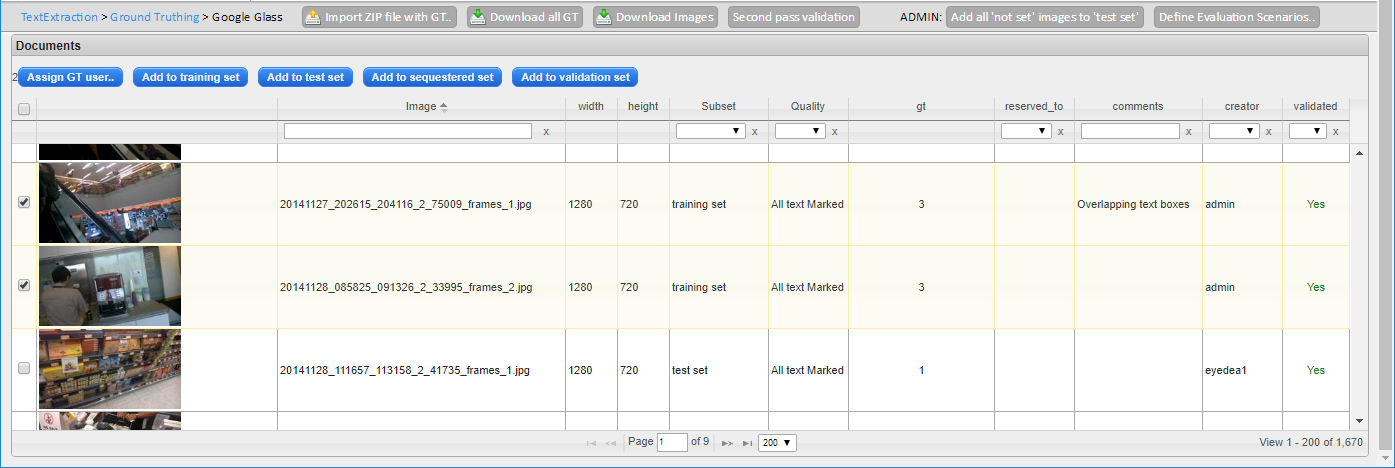}
	\caption{A reduced screenshot of the ground truth management tool.}
	\label{fig:GT_Management}
\end{figure}

Figure~\ref{fig:GT_Management} shows the annotation management dashboard, which presents a searchable list of images along with their status and other metadata to the manager of the annotation operation. The annotation manager can make use of this information to provide feedback and ensure consistency of the annotation process. The dashboard allows keeping track of the overall progress, responding to specific comments that annotators make, requesting a revision of the annotations and assigning quality ratings to images, while it provides version control and coordination mechanisms between annotators.

The dashboard allows either assigning images to specific annotators, or letting annotators select the images to work on. Assigning specific images to annotators is useful when we want to ensure that no individual annotator has access to the whole dataset. Annotators can reserve images for a period of time to continue in various sessions. The same interface allows assigning images to the different subsets (training, validation, public and sequestered test) that are then used for defining evaluation scenarios.

Using the RRC Annotation and Evaluation Platform has evolved over time to support image annotation at different granularities, making it possible to generate annotations from the pixel level to text lines. The RRC Platform stores annotations internally as a hierarchical tree using a combination of XML files for metadata and transcription information and image files for pixel level annotations.

\begin{figure*}
	\centering
	\includegraphics[width=\linewidth]{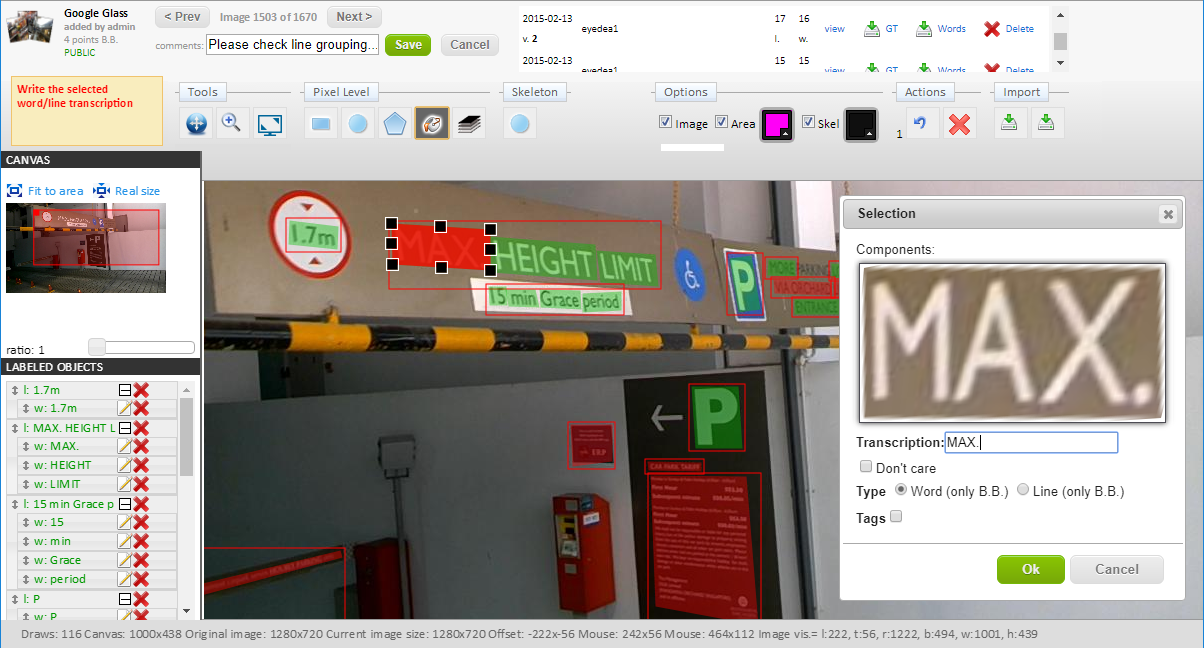}
	\caption{Screenshot of one of the Web-based annotation tools. When defining 4-point quadrilateral bounding boxes annotators are shown a real time preview of a rectified version of the region being defined.}
	\label{fig:GTTool}
\end{figure*}

A screenshot of one of the Web-based annotation tools can be seen in Figure~\ref{fig:GTTool}. The hierarchy of textual content and the defined text parts is displayed on the left-hand side of the interface. In the example shown, annotations are defined at the word level (axis oriented or 4-point quadrilaterals) and grouped together to form text lines. Alternatively, annotations can be made at different granularities: pixel-level, atoms, characters, words and text blocks are supported.

A number of tools are provided to ensure consistency and quality during the annotation process, we detail two of them here.

\subsubsection{Perspective text annotation}
As new more demanding challenges were introduced over time, the need to deal with text with high perspective distortions arose (e.g. the Incidental Text challenge~\cite{karatzas2015icdar}). Consequently, instead of axis-oriented bounding boxes, we introduced the possibility to define 4-point quadrilateral bounding boxes around words or text lines.

When 4-point quadrilateral bounding boxes are defined around text with perspective distortion, it is inherently difficult for annotators to agree on what is a good annotation and provide meaningful instructions. To ensure consistency, we introduced a real time preview of a rectified view of the region being annotated. Annotators are then required to adjust the quadrilateral so that the \textit{rectified} word appears straight (see inlet in Figure~\ref{fig:GTTool}). We observed that this process improves substantially the consistency between different annotators, and speeds up annotation.


\subsubsection{Deciding what is unreadable}
All annotated elements, apart from their transcription, can have any number of custom defined associated metadata like script information, quality metrics etc. A special type is reserved for text that should be excluded from the evaluation process, and is thus marked as \textit{do not care}
. Depending on the challenge, such cases can include text which is partially cut, low resolution text, text in scripts other than the ones the challenge focuses on, or indeed any other text that the annotator deems as unreadable text.

\begin{figure}
	\centering
	\includegraphics[width=\linewidth]{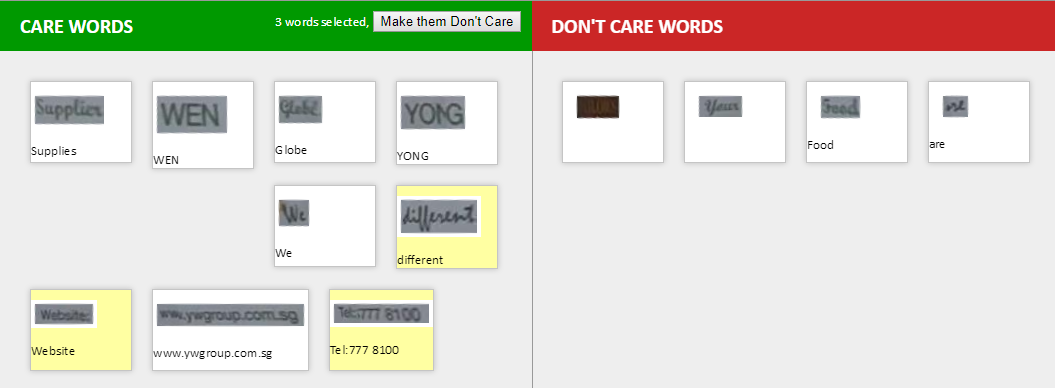}
	\caption{Drag-and-drop interface for validating do not care words at the image level.}
	\label{fig:DoNotCareDragDrop}
\end{figure}

\begin{figure}
	\centering
	\includegraphics[width=\linewidth]{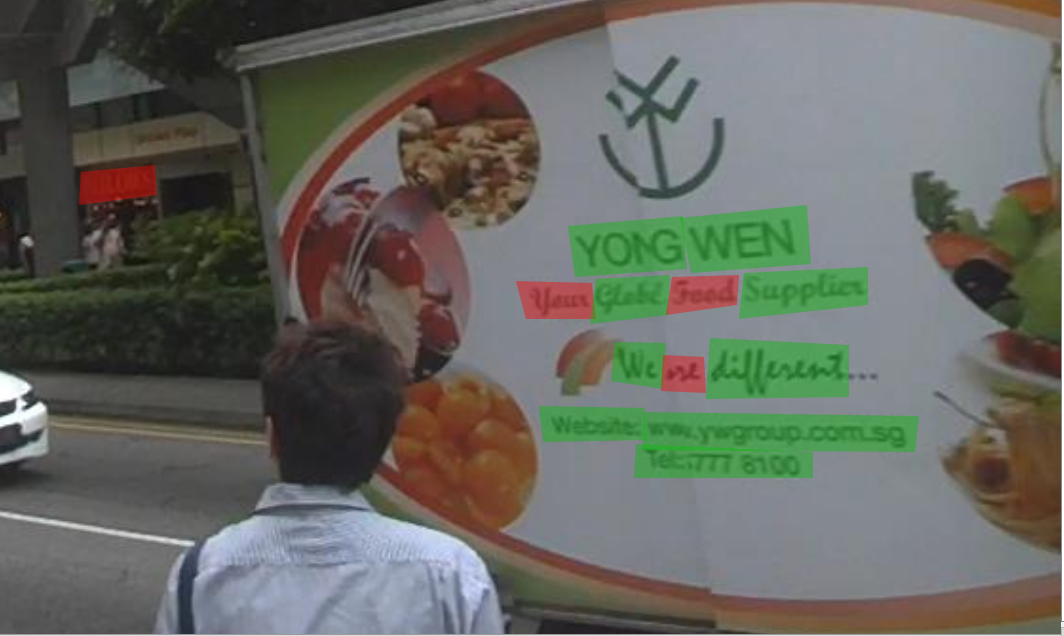}
	\caption{\textit{Do not care} regions appear in red, normal regions appear in green. In the example, words have gone through a second-stage verifications where their readability was judged individually to eliminate any annotation bias introduced by contextual information (e.g. words that can be guessed to say ``food'' due to the visual context, or ``are'' due to textual context were judged as unreadable when seen individually).}
	\label{fig:DoNotCareContext}
\end{figure}

Judging whether a text instance is unreadable and should be marked as \textit{do not care} is challenging, and in some cases similar text is treated differently by different annotators. At the same time, there are cases where reading is assisted by the textual or visual context (e.g. if the words on the left and right are readable the middle word can be easily guessed), and annotators have trouble deciding whether such text should be actually marked as \textit{do not care} or not. To reduce subjective judgements we have implemented various verification processes. First, annotators can explore all words of a particular image out-of-context grouped according to their status, through an interface that allows dragging words between the \textit{do not care} and \textit{care} sides (see Figure~\ref{fig:DoNotCareDragDrop}). This ensures per-image consistency of the annotations. At a later stage, a second-pass verification process is introduced through an interface that displays words of the whole dataset individually, out of context and in random order to be verified on their own. This has been shown to eliminate the inherent bias of annotators to use surrounding textual or visual context to guess the transcription (see Figure~\ref{fig:DoNotCareContext}).

\subsection{Definition of Research Tasks}
The competition is structured in challenges and research tasks. Challenges correspond to specific datasets, representing different domains such as born-digital images, real-scene images or videos obtained in different scenarios, etc, while research tasks (e.g. text localisation, text recognition, script identification, end-to-end reading, etc) are defined for each of these challenges. The units of evaluation are the research tasks. Two key aspects define the process of evaluation of a research method over a particular research task: the data and the evaluation script to be used.

As mentioned before, datasets can be fully dealt with within the framework. It is nevertheless quite often nowadays that datasets and annotations have been obtained in various different or complementary ways (e.g. crowd-sourcing).
The RRC platform supports defining a research task based on either internally curated data or directly linking to externally provided annotations.

The evaluation scripts are the key elements through which the submitted files (e.g. word detections produced by a method), are processed and compared against the ground truth annotations producing evaluation results. Evaluation results are produced in terms of overall metrics over the whole dataset, but can be also optionally produced at a per-sample level, which enables further analysis and visualisation of results.

In addition, the evaluation scripts perform other auxiliary functions such as validating an input file against the expected format (a process used by the Web portal to early reject submissions and inform authors of problems). All evaluation scripts are currently written in Python
. Apart from the on-line evaluation interface, all evaluation scripts are available to download through the RRC web portal, and can be used from the command prompt.


A graphical user interface accessible through the RRC portal permits linking together the different aspects that comprise a research task (data files, evaluation and visualisation scripts), and generates the submission forms and results visualisation pages of the on-line competition portal, as well as stand-alone versions of the Web interfaces that can be used off-line (see next section).  
Note that multiple evaluations can be defined in parallel for the same task (e.g. a text localisation task can be evaluated using an Intersection-over-Union scheme or a more classic DetEval type evaluation).


Evaluation is then managed on the server side by launching separate services for each evaluation task that needs to be performed. The services automatically look for new submissions to evaluate and produce results in designated places. This permits us to launch a variable number of instances of the evaluation service dedicated to a specific task, on the same or different servers, resulting in a neat mechanism of achieving balancing and horizontal scalability.

\subsection{Evaluation and Visualisation of Results}
\begin{figure}
	\centering
	\includegraphics[width=\linewidth]{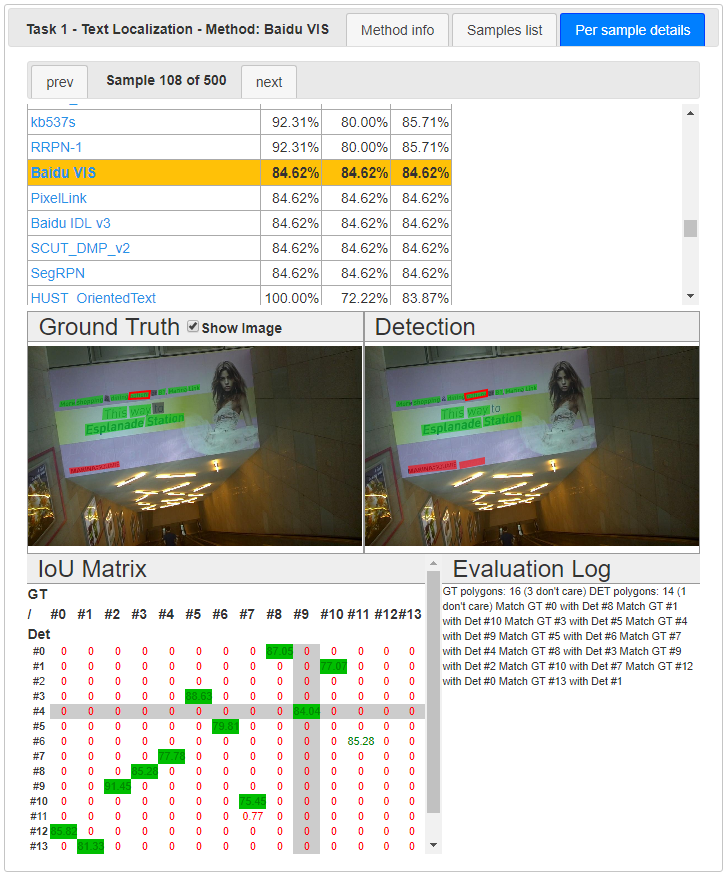}
	\caption{Per-image results interface for text localisation.}
	\label{fig:PerImageResults}
\end{figure}

The on-line portal permits users to upload the results of their methods against a public validation / test dataset and obtain evaluation results on-line. Apart from ranked tables of quantitative results of submitted methods, users can explore per sample visualisations of their results along with insights about the intermediate evaluation steps, as seen in Figure~\ref{fig:PerImageResults}. Through the same interface users can hot-swap between different submitted methods to easily compare behaviours.

\begin{figure}
	\centering
	\includegraphics[width=\linewidth]{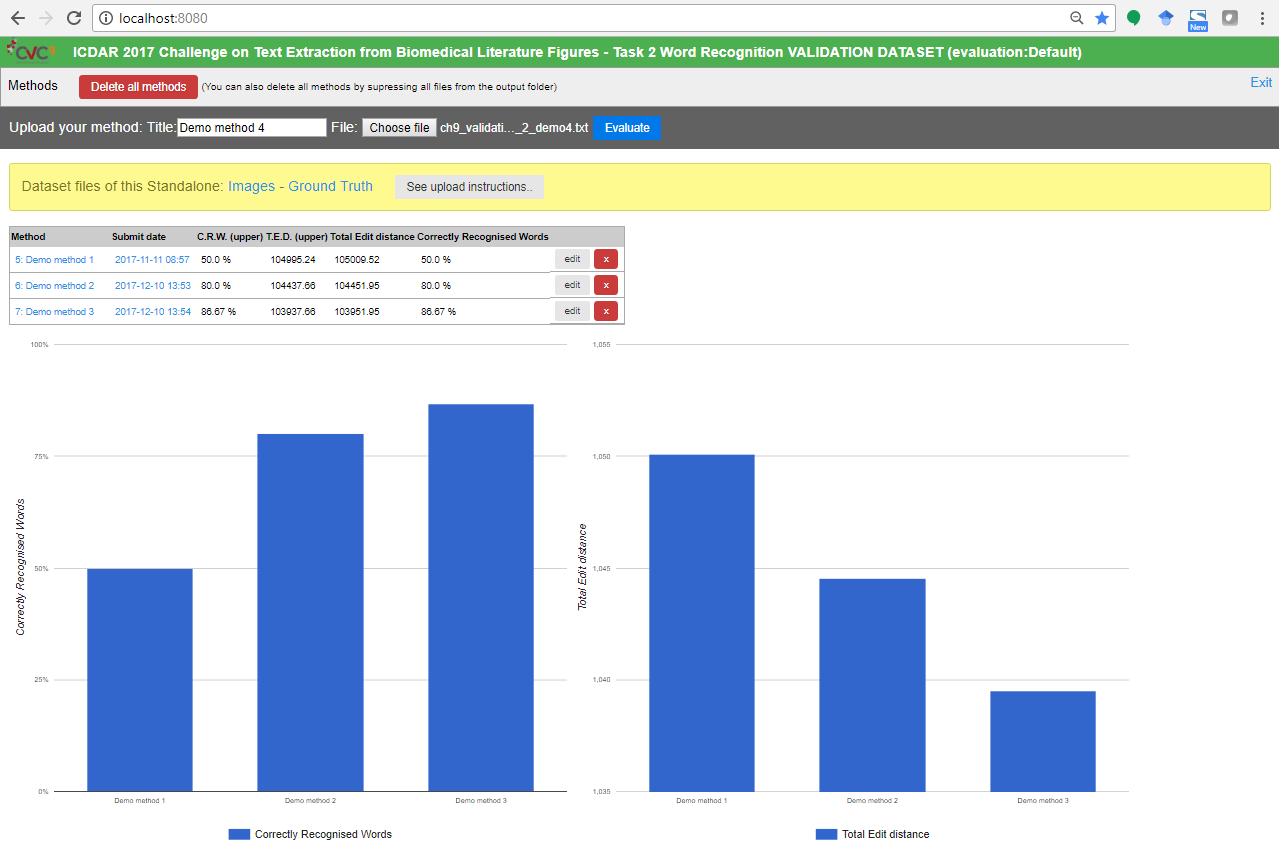}
	\caption{View of the home page of the standalone Web interface, running locally.}
	\label{fig:StandAlone}
\end{figure}

All the evaluation and visualisation functionality can also be used off-line. As part of the research task deployment process described before, a downloadable version of a mini-Web portal is also produced, which packs together a standalone Web server along with all data files and evaluation scripts necessary to reproduce the evaluation and visualisation functionality off-line. Figure~\ref{fig:StandAlone} shows the home page of the standalone server for the text localisation task of the DeText challenge, running on the local host.

\section{Conclusion}
The RRC Annotation and Evaluation Platform is the backbone of the Robust Reading Competition's on-line portal. It comprises a number of tools and interfaces that are available for research use. The goal of this paper is to raise awareness about the availability of these tools, as well as to share insights and best-practices based on our experience with organising the RRC over the past 7 years.

The evaluation and visualisation functionalities of the portal are available on-line\footnote{\url{http://rrc.cvc.uab.es}}, and currently being used by thousands of researchers. In parallel, the whole Web portal functionality along with evaluation scripts is available to download and use off-line through standalone implementations. Access to the latest data management and annotation interfaces is possible for research purposes through the RRC portal by contacting the authors requesting access, while a limited functionality (2014 version) of the annotation tools is available to download and use off-line\footnote{\url{http://www.cvc.uab.es/apep/}}.

\begin{figure}
	\centering
	\includegraphics[width=\linewidth]{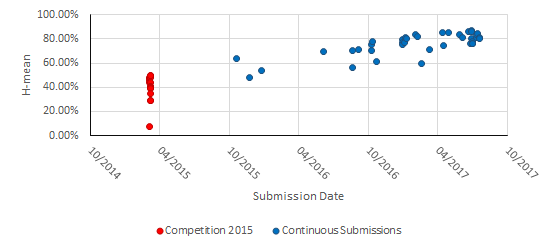}
	\caption{Evolution of performance for the text localisation task on the incidental scene text challenge over time.}
	\label{fig:SoAEvolution}
\end{figure}

We are continuously working on new functionality. The next key changes will be related to methods' metadata and the versioning system of the platform. One of our ambitions is to be able to produce meaningful real-time insights on the evolution of the state of the art, based on the information collected over time (see for example Figure~\ref{fig:SoAEvolution}). We hope that the tools currently offered (on-line private submissions and off-line standalone Web interface) should already help individual researchers to track their progress.


\section*{Acknowledgements}
This work is supported by Spanish projects \mbox{TIN2014-52072-P}, \mbox{TIN2017-89779-P} and the CERCA Programme / Generalitat de Catalunya.



\newcommand{\BIBdecl}{\setlength{\itemsep}{0.25 em}}
\bibliographystyle{IEEEtran}
\bibliography{paper}
%
%
%
%

\end{document}